\documentclass{Interspeech}

\interspeechcameraready

\usepackage{multirow}
\title{Benchmarking and Confidence Evaluation of LALMs For Temporal Reasoning}

\author[affiliation={1}]{Debarpan}{Bhattacharya*}
\author[affiliation={1}]{Apoorva}{Kulkarni*}
\author[affiliation={1}]{Sriram}{Ganapathy}
\def\x{{\mathbf x}}

\affiliation{Electrical Engineering}{Indian Institute of Science}{India}
\email{debarpanb@iisc.ac.in, apoorvakulkarni2001@gmail.com, sriramg@iisc.ac.in}

\usepackage{comment}
\usepackage{multirow}
\usepackage{arydshln}
\usepackage{xcolor}
\begin{document}
\maketitle

\begin{abstract}
The popular success of text-based large language models (LLM) has streamlined the attention of the multimodal community to combine other modalities like vision and audio along with text to achieve similar multimodal capabilities. 
In this quest, large audio language models (LALMs)  have to be evaluated on reasoning related tasks which are different from traditional classification or generation tasks. 
Towards this goal, we propose a novel dataset called temporal reasoning evaluation of audio (TREA).
We benchmark open-source LALMs and observe that they are consistently  behind human capabilities on the tasks in the TREA dataset. 
While evaluating LALMs, we also propose an uncertainty metric, which computes the invariance of the model to semantically identical perturbations of the input.
Our analysis shows that the accuracy and uncertainty metrics are not necessarily correlated and thus, points to a need for wholesome evaluation of LALMs for high-stakes applications. The dataset and code are made public.\footnote{\noindent\url{https://github.com/iiscleap/Audio-LLM-benchmarking-uncertainty}\\
* equal contribution.\\
Accepted in INTERSPEECH, 2025, Rotterdam, The Netherlands.
}
\end{abstract}
\section{Introduction}
\label{sec:introduction}
The development of text-based large language models (LLM) like GPT~\cite{achiam2023gpt}, LLAMA~\cite{touvron2023llama}, Gemini~\cite{team2023gemini}, Mistral~\cite{jiang2023mistral}, etc., have pushed the envelope on text processing capabilities, driven by unsupervised pre-training on humongous amount of text corpus crawled from the internet. These models have made tremendous advances in achieving human and super human level performances on tasks like creative writing, mathematics and coding, question answering and texual reasoning~\cite{hendrycksmeasuring}. 
However, in the quest towards artificial general intelligence (AGI), it is found that current 
multi-modal skills are inferior to the text-based capabilities  \cite{morris2023levels}.
The key challenge in achieving this ability is the requirement of multi-modal semantic alignment \cite{zhang2024mm}.
In some of the prominent open-source models, the multimodal alignment is addressed by fusing the text embeddings with the  modality specific vision/audio encoder, and with the light-weight adaptation of text-only models. Examples of such models include large vision language models (LVLMs)(e.g. LLAVA~\cite{liu2024visual}, LLAMA 3V~\cite{hanoona2024LLaVA++} and large audio language models (LALMs) (e.g. Qwen-audio~\cite{chu2023qwen}, SALMONN~\cite{tang2023salmonn}, WavLLM~\cite{hu-etal-2024-wavllm}).

One of the key steps in establishing the efficacy and skill-set of LLMs is the development of systematic evaluation protocols. 
The text-based LLMs have seen substantial progress in benchmarking and standardization (For example, chatbot arena - \cite{chiang2024chatbot}, MMLU~\cite{hendrycksmeasuring}).
A detailed survey of LLM evaluation efforts is available in \cite{chang2024survey}. Similar efforts in  LVLMs have also matured \cite{xu2024lvlm} and have in-turn propelled the advancement of these models. However, recent benchmarking efforts~\cite{fu2024blink} have reflected that their multimodal performance wane specially in cross modal understanding and reasoning tasks.

Although a handful of studies   probe speech/audio foundation models for standard tasks like automatic speech recognition (ASR), speech synthesis, emotion recognition, speaker identification, and other paralinguistic tasks~\cite{hu-etal-2024-wavllm, gao2025ttslow, wang2024can, cui2024recent, sakshi2024mmau, tang2023salmonn, waheed2024speech}, there are limited studies that report domain specific evaluation of LALMs in audio reasoning and understanding tasks. Such focused benchmarking remains paramount as they help to unveil whether the models comprehend the occurrence of different events in an audio, distinguish them, and understand their mutual correspondence.

While LALM model development is active, the data, the skills as well as the performance metrics to analyze LALMs for fine-grained tasks like temporal reasoning require significant research undertakings to understand the modeling gaps.

In this paper, we contribute a novel dataset for benchmarking the temporal reasoning capabilities of LALMs and show that the current LALMs are significantly lacking in such skills. We highlight that the lack of such skills indicate the inability of models to understand fine-grained temporal information and in extracting this information for a question-answering task.
Further, the paper also proposes methods for measuring the uncertainty of LALMs that focus on the resilience of the model to test-time perturbations that are semantically grounded. 

\section{Related Work and Contributions}
\label{sec:recent_wors_contributions}
\textbf{Large Audio Language Models (LALMs)}: The early work on extending text-based LLMs to audio domain can be traced to efforts like Audio-PALM \cite{rubenstein2023audiopalm}, CLAP \cite{elizalde2023clap} and Speech-GPT \cite{zhang2023speechgpt}. With the popularization of the open-source LLaMa based models \cite{touvron2023llama} and with the availability of open-source speech recognition systems (Whisper - \cite{radford2023robust}), LALMs were developed using speech/audio encoders (Whisper-based) and LLaMa back-end. The pre-trained LLaMa LLMs were fine-tuned for speech/audio tasks using low-rank adapters \cite{hu2021lora}. Examples of such models include Qwen-audio~\cite{chu2023qwen}, SALMONN~\cite{tang2023salmonn}, and WavLLM~\cite{hu-etal-2024-wavllm}. In this work, we explore these three models for our evaluations.
\begin{figure}[t!]
    \centering
    \includegraphics[width=1.0\linewidth]{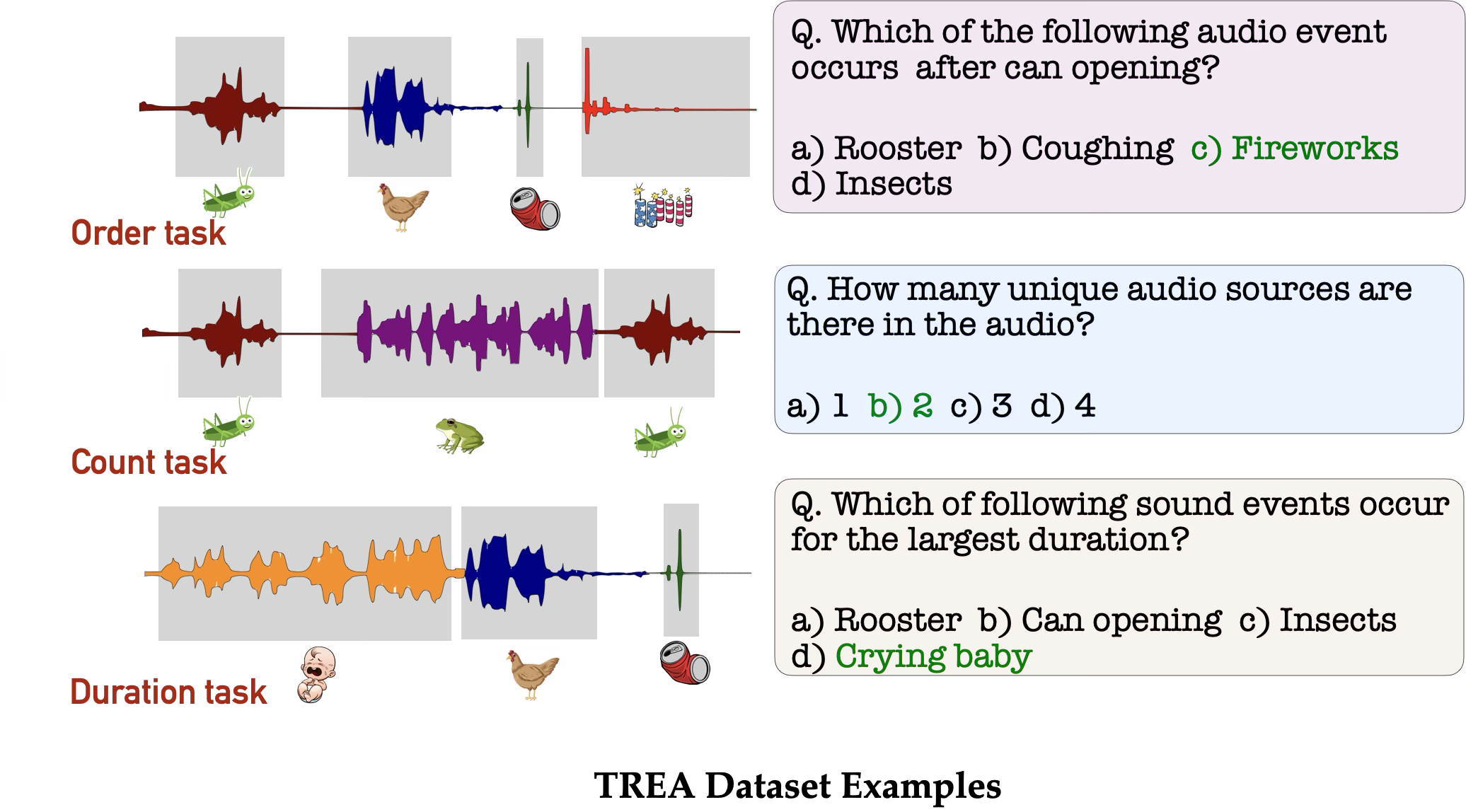}
    \caption{Examples from TREA dataset illustrating the sub-tasks in event order, count and duration based MCQA.} 
    \vspace{-0.2in}
    \label{fig:audio-tasks-dataset}
\end{figure}
\noindent \textbf{Benchmarking LALMs}: A recent effort to evaluate LALMs for multi-audio tasks was undertaken by Chen et. al. \cite{chen2024beyond}. 
In the same vein, Sakshi et. al. \cite{sakshi2024mmau} developed a large scale benchmark for audio understanding, information extraction and question-answering. Another benchmarking effort was pursued by Yang et. al. \cite{yang2024air}. In this evaluation, $19$ sub-tasks were probed which involved speech, music and sound classification tasks.  In spite of these efforts, temporal reasoning capabilities are largely unexplored. However, in the  image domain, visual spatial reasoning has been probed elaborately through works like \cite{wang2025picture} and with novel datasets like BLINK \cite{fu2024blink}.  In this paper, we propose a dedicated dataset for probing the temporal reasoning capabilities of LALMs.

\noindent \textbf{Evaluating Models Beyond Accuracy}: The performance of classification style tasks (which also include multi-choice question-answering (MCQA)) has primarily been based on accuracy or F1 score. Beyond accuracy, the two primary directions of model evaluation include calibration and confidence estimation \cite{geng2024survey}. 
Calibration refers to the correlation between the model's performance (accuracy) and the posterior-probability score \cite{jia2021scaling}. A well-calibrated model generates more accurate answers to MCQA when the token probability of the answer is higher. 
However, uncertainty estimation measures how confident the model is for a given input around the local neighbourhood of the given input. 
In natural language generation (NLG), the uncertainty is measured using various output samplings that are semantically equivalent \cite{kuhn2023semantic}. However, in an MCQA setting, it is more appropriate to measure the epistamic uncertainty \cite{shorinwa2024survey} which measures the resilience of the model to semantically invariant perturbations of the input. 
In deep learning based image classification setting, a similar uncertainty measure was proposed \cite{bahat2020classification}. In this paper, we adapt this framework for audio-based uncertainty estimation of LALMs on the temporal reasoning tasks.\\
\noindent\textbf{Key Contributions :}
\begin{itemize}
    \item We propose an open-source audio dataset of MCQAs specifically designed for probing temporal reasoning abilities, within the sub-task categories of  event ordering, counting, and   duration reasoning (\textbf{Section~\ref{sec:dataset}}).
    \item We perform task-wise benchmarking of prominent LALMs and highlight the shortcomings of these models on temporal reasoning  tasks. Also, we show that the models perform differently on different reasoning sub-tasks, emphasizing the need for fine-grained evaluations (\textbf{Section~\ref{sec:performance}}).
    \item We propose an uncertainty measure using test-time data augmentation and report this metric along with calibration errors for the LALMs. We show that the performance and uncertainty measures are not correlated. Both being important, our study calls for benchmarking uncertainty measures, besides the performance evaluations(\textbf{Section~\ref{sec:metrics_uncertainty}}).
\end{itemize}

\begin{table}[t!]
\centering
\resizebox{1.0\linewidth}{!}{
\def\arraystretch{1.4}
\begin{tabular}{lllll}
\toprule[0.75pt]
\textbf{Name} & \textbf{Size (Reasoning)}& \textbf{Tasks} & \textbf{Target Resp.} & \textbf{Comment} \\ 
\hline
CompA-R~\cite{ghosh2024gama} & $1.5k (1.5k)$ & Ordering & True/False & Binary QA \\ 
MusicBench~\cite{melechovsky2023mustango} & $0.4k (0)$ & Captioning & NLG & Music Desc. \\ 
MuChoMusic~\cite{weck2024muchomusic} & $1.2k(0.4k)$ & Genre,  Instruments  & NLG &  Reasoning \\ 
AudioBench~\cite{wang2024audiobench} &  $100k (0)$ & Speech, Audio & All & Capt., QA, Class. \\ 
AIR-Bench~\cite{yang2024air} & $20k (0.8k)$ & Audio, Music, Speech   & All & Capt., QA, Class. \\ 
MMAU~\cite{sakshi2024mmau} & $1k (0.7k)$ & Audio, Music & MCQA &   Reasoning, QA.  \\\hline
\textbf{TREA (ours)} & $0.6k (0.6k)$ & Audio &  MCQA & Order, Count, Dur. \\ \bottomrule[0.75pt]
\end{tabular}}
\vspace{0.05in}
\caption{Comparison of various audio benchmarks for evaluating LALMs with ours. Here, NLG means natural language generation.}
\label{tab:dataset-comparison}
\vspace{-0.3in}
\end{table}
\section{Dataset Design}
\label{sec:dataset}
We design a dataset to benchmark temporal reasoning tasks in audio, termed temporal reasoning evaluation of audio (TREA). 
It probes the models on three sub-tasks -  event duration (TREA-D), ordering (TREA-O) and counting (TREA-C). 
The dataset size is $600$ samples, with each sub-task containing $200$ samples.

The dataset is derived from the Environmental Sound Classification dataset (ESC-50) \cite{piczak2015esc}, having audio recordings from $50$ different environmental sound classes. Each recording is of $5$ sec duration in the ESC-50 dataset.  All the audio files in the TREA dataset are obtained by combining multiple ESC-50 audio recordings. To compute effective duration of the ESC recordings, an activity detection using a simple energy based threshold is used to discard silent regions of the audio. Examples from the TREA dataset are shown in Fig.~\ref{fig:audio-tasks-dataset}. 
\begin{table*}[t!]
\centering
\resizebox{0.9\linewidth}{!}{
\bgroup
\def\arraystretch{1.2}
\begin{tabular}{l|ccl|lcl|ccc|cccccc}
\toprule[0.75pt]
\toprule[0.75pt]
\multicolumn{1}{c|}{\multirow{2}{*}{\textbf{Audio-LLM(s)}}} &  & \multirow{2}{*}{\textbf{Size}} &  &  & \multirow{2}{*}{\textbf{Baselines}} &  & \multicolumn{1}{l}{} & \textbf{MMAU} & \multicolumn{1}{l|}{} & \textbf{} & \multicolumn{5}{c}{\textbf{TREA}} \\ \cline{9-9} \cline{12-16} 
\multicolumn{1}{c|}{} &  &  &  &  &  &  & \multicolumn{1}{l}{} & \textbf{Temp. Reason.} & \multicolumn{1}{l|}{} & \multicolumn{1}{l}{} & \textbf{Dur. } & \textbf{} & \textbf{Order } & \multicolumn{1}{l}{\textbf{}} & \textbf{Count } \\ \hline
Random &  & - &  &  & - &  &  & - &  &  & 27.5 &  & 23.5 &  & 26.5\\\hline
Human &  & - &  &  & - &  &  & - &  &  & 92.2 &  & 98.9 &  & 81.1  \\\hline
\multirow{4}{*}{Qwen2-Audio~\cite{chu2023qwen}} &  & \multirow{4}{*}{7B} &  &  & Vanilla &   &  & 33.3  & &  & 38.5 &  & 61.0 &  & 31.0 \\
 &  &  &  &  & CoT &  &  & 31.3 &  &  & 44.5 &  & 49.0 &  & 16.5 \\
 &  &  &  &  & Explanation &  &  & 37.5 &  &  & 39.5 &  & 59.0 &  & 23.5 \\
 \cmidrule{2-16}
&  & 7B + 70B &  &  & Desc. + LLM-QA &  &  & 41.7 &  &  & 44.5 &  & 51.0 &  & 21.0 \\ \midrule 
\multirow{4}{*}{SALMONN~\cite{tang2023salmonn}} &  &\multirow{4}{*}{13B}  &  &  & Vanilla &  &  & 41.7 &  &  & 42.0 &  & 34.0 &  & 20.5 \\
 &  &  &  &  & CoT &  &  & 37.5 &  &  & 34.0 &  & 46.5 &  & 23.0 \\
 &  &  &  &  & Explanation &  &  & 35.4 &  &  & 37.5 &  & 51.5 &  &  27.5\\ 
 \cmidrule{2-16}
 &  & 13B + 70B &  &  & Desc. + LLM-QA &  &  & \textbf{47.9} &  &  & \textbf{47.0 }&  & \textbf{69.0} &  & \textbf{45.5} \\ 
 \midrule 
\multirow{4}{*}{WavLLM~\cite{hu-etal-2024-wavllm}} &  & \multirow{4}{*}{7B} &  &  & Vanilla &  &  & 29.2 &  &  & 29.5 &  & 32.5 &  & 20.5 \\
 &  & &  &  & CoT &  &  & 22.9 &  &  & 32.0 &  & 28.5 &  & 24.0 \\
 &  &  &  &  & Explanation &  &  & 23.0 &  &  & 24.0 &  & 32.5 &  & 33.0 \\ \cmidrule{2-16}  
 &  & 7B + 70B &  &  & Desc. + LLM-QA &  &  & 22.9 &  &  & 22.5 &  & 20.5 &  & 11.5 \\ 
 
 \hline   
\end{tabular}
\egroup
}
\vspace{0.05in}
\caption{Performance (Accuracy \%) of various LALMs  on temporal reasoning tasks in MMAU and TREA dataset. For description + LLM-QA, the LALM provides the audio description and an inference call is made to \texttt{LLaMa-3.3-70B} \cite{touvron2023llama} model to perform text MCQA using the audio captions.}
\label{tab:perf_benchmark}
\vspace{-0.3in}
\end{table*}
\subsection{TREA-D: Audio event duration}

This subset of the TREA data contains audio files related to the event duration MCQA. 
For example, an MCQA based on event duration is - (\textit{which of the following sound events correspond to the longest duration?}). 
During the combination of the audio recordings from ESC-50, it is ensured  that the duration of the largest (shortest)  event is significantly more (less) than the duration of  other events in the combined audio file. This ensures that there is no ambiguity in the ground-truth labels.
 
\subsection{TREA-O: Audio event ordering}
This data subset focuses on probing if the LALM model understands  the temporal order of the audio-events. An example of an MCQA is - (\textit{which sound event occurred after ``can opening'' sound?}). To avoid any ambiguity, it is ensured that none of the audio events  repeat in the same recording. 
\subsection{TREA-C: Audio event counting}
This subset consists of audio recordings that contain repetitive events of the same event class. The model is probed with MCQA questions like -  (\textit{how many unique sound sources are present in the audio file?}).
\subsection{Comparison with other datasets}
The comparison of the proposed dataset  with other audio evaluation benchmarks is given in Table~\ref{tab:dataset-comparison}. As seen here, the other datasets do not largely focus on the temporal reasoning tasks in audio. 
The only exception is the MMAU dataset, where audio reasoning tasks are included. However, the public release of MMAU has only a small number of temporal reasoning samples ($48$ samples). Further, the MMAU dataset has only one audio event per sample.
The proposed TREA dataset attempts to bridge these gaps and aims to provide insights about the fine-grained audio reasoning capabilities of the current LALMs.

\section{Benchmarking and analysis}
\label{sec:performance}
 We consider audio temporal reasoning subset of MMAU dataset~\cite{sakshi2024mmau} ($48$ samples), and the proposed TREA dataset (total $600$ samples spread across $3$ fine grained tasks). The LALMs are evaluated in zero-shot manner and the performance (accuracy ($\%$)) is  reported in Table~\ref{tab:perf_benchmark}. The top row (\textit{random}) denotes an experiment where one of the $4$ options were chosen randomly for all the questions. The second row (\textit{human}) represents the human performance on these tasks, where $6$ participants performed the task on a smaller subset ($45$ questions, $15$ for each sub-task) from the TREA dataset.
 We perform the LALM evaluations in multiple settings,

\begin{itemize}
    \item \textbf{Vanilla}: This is basic version where the LALM is prompted to respond to the MCQA with response restricted to the answer choice among the four options. 
    \item \textbf{Chain-of-thought (CoT)}: In this setting, the LALM is prompted to process the question, and then think step-by-step leading to the final answer choice~\cite{wei2022chain}. 
    \item \textbf{Explanation}: In this setting, the LALM is asked to provide the best answer choice and then provide an explanation as to why the answer choice was made.  
    \item \textbf{Audio description + LLM-QA}: Here, LALM is used in generative fashion. It is prompted with only the audio data and asked to provide a detailed audio caption. Further, the caption along with the textual MCQ is given to a text-based \texttt{LLaMa-3.3-70B-Instruct} model to choose the right option. 
\end{itemize}

\noindent The key takeaways from this evaluation (Table~\ref{tab:perf_benchmark}) are:
 \begin{itemize}
 \item We observe that the performance on audio temporal reasoning tasks are poor for all the LALMs (the best performance is $<50$\% on two out of the three TREA tasks), highlighting their lack of understanding of audio events in the temporal domain. Note that, the human performance on $2$ tasks is $>90\%$ and $1$ task is $>80\%$, illustrating a large performance gap with LALMs.  
 \item The performance in reasoning sub-tasks is largely varied- it is moderate in ordering and poor in duration and counting tasks. This shows the importance of benchmarking with respect to fine-grained reasoning sub tasks.
     \item In vanilla experiments, Qwen performs the best in order and count task of TREA dataset, whereas SALMONN performs the best in duration task of TREA dataset and in MMAU.
     \item The benefits of prompt based reasoning strategies like CoT and explanations are not very clear unlike the case in text-based LLMs, where CoT has shown clear advantages~\cite{wei2022chain}.
     \item Using LALMs in generative mode to generate audio captions and then using a separate LLM (LlaMa 3.3) to select right answer choice from the audio caption produces interesting results. SALMONN generated captions result in $67.1\%$ mean relative improvement in TREA dataset and $12.9\%$ relative improvement in MMAU dataset. However, performances for Qwen and WavLLM are worse than thosefrom SALMONN.
     \item We hypothesize that, the multimodal discriminative ability of SALMONN is poor, and hence it fails to achieve high accuracy in spite of having high quality captions generated by SALMONN.
 \end{itemize}
\begin{table*}[t!]
\centering
\resizebox{1.0\linewidth}{!}{
\def\arraystretch{1.2}
\begin{tabular}{l|ccc|ccc|ccc|ccc}
\toprule[0.75pt]
\toprule[0.75pt]
\multirow{2}{*}{\textbf{Audio-LLM(s)}} & \multicolumn{3}{c|}{\textbf{Duration Task}} & \multicolumn{3}{c|}{\textbf{Ordering Task}} & \multicolumn{3}{c|}{\textbf{Counting Task}} & \multicolumn{3}{c}{\textbf{Combined}} \\ 
 \cline{2-13} 
   & \textbf{Acc.} ($\uparrow$) & \textbf{ECE } ($\downarrow$) & \textbf{EUE } ($\downarrow$) & \textbf{Acc.} ($\uparrow$) & \textbf{ECE} ($\downarrow$) & \textbf{EUE} ($\downarrow$) & \textbf{Acc.} ($\uparrow$) & \textbf{ECE} ($\downarrow$) & \textbf{EUE} ($\downarrow$) & \textbf{Acc.} ($\uparrow$) & \textbf{ECE} ($\downarrow$) & \textbf{EUE} ($\downarrow$) \\ 
\hline
Qwen2-Audio-Instruct-7B & 48.7 & 27.5 & 31.1 & 49.3 & 27.1 & \textbf{23.0} & 16.6 & 60.27 & 24.4 & 38.2 & 38.3 & 26.2 \\ 
SALMONN-13B & 46.2 & \textbf{13.9} & \textbf{24.3} & 41.0 & \textbf{12.7} & 33.2 & 17.3 & \textbf{42.2} & \textbf{7.0} & 34.9 & \textbf{18.22} & \textbf{21.5} \\ 
\hdashline
SALMONN+LLaMa-83B & \textbf{52.4} & 43.8 & 41.9 & \textbf{63.8} & 31.7 & 26.7 & \textbf{51.7} & 47.6 & 53.2  & \textbf{55.9} & 41.0 & 40.6 \\ 
\bottomrule[0.75pt]
\end{tabular}
}
\vspace{0.05in}
\caption{Benchmarking LALMs: Accuracy on the perturbed samples, expected calibration error (ECE), and expected uncertainty (EUE). All values are in \%. }
\label{tab:calibration-and-uncertainty}
\vspace{-0.3in} 
\end{table*}

\section{Metrics Beyond Accuracy}
\label{sec:metrics_uncertainty}
\subsection{Test-time Uncertainty Measure}
We propose to measure the uncertainty in the decision making for a test sample using data perturbations. 
The perturbations are generated in such a way that the semantic content of the perturbed sample is identical to the original test sample (unaltered ground truth label)
The analogy in image classification is the perturbations created using the same object with different orientations, contrast and zoom \cite{bahat2020classification}, without altering the object class present in the image. 
The steps involved in computation of the uncertainty measure for LALMs are the following. 

\subsubsection{Data augmentation}
\begin{figure}[t!]
    \centering
    \includegraphics[width=0.9\linewidth]{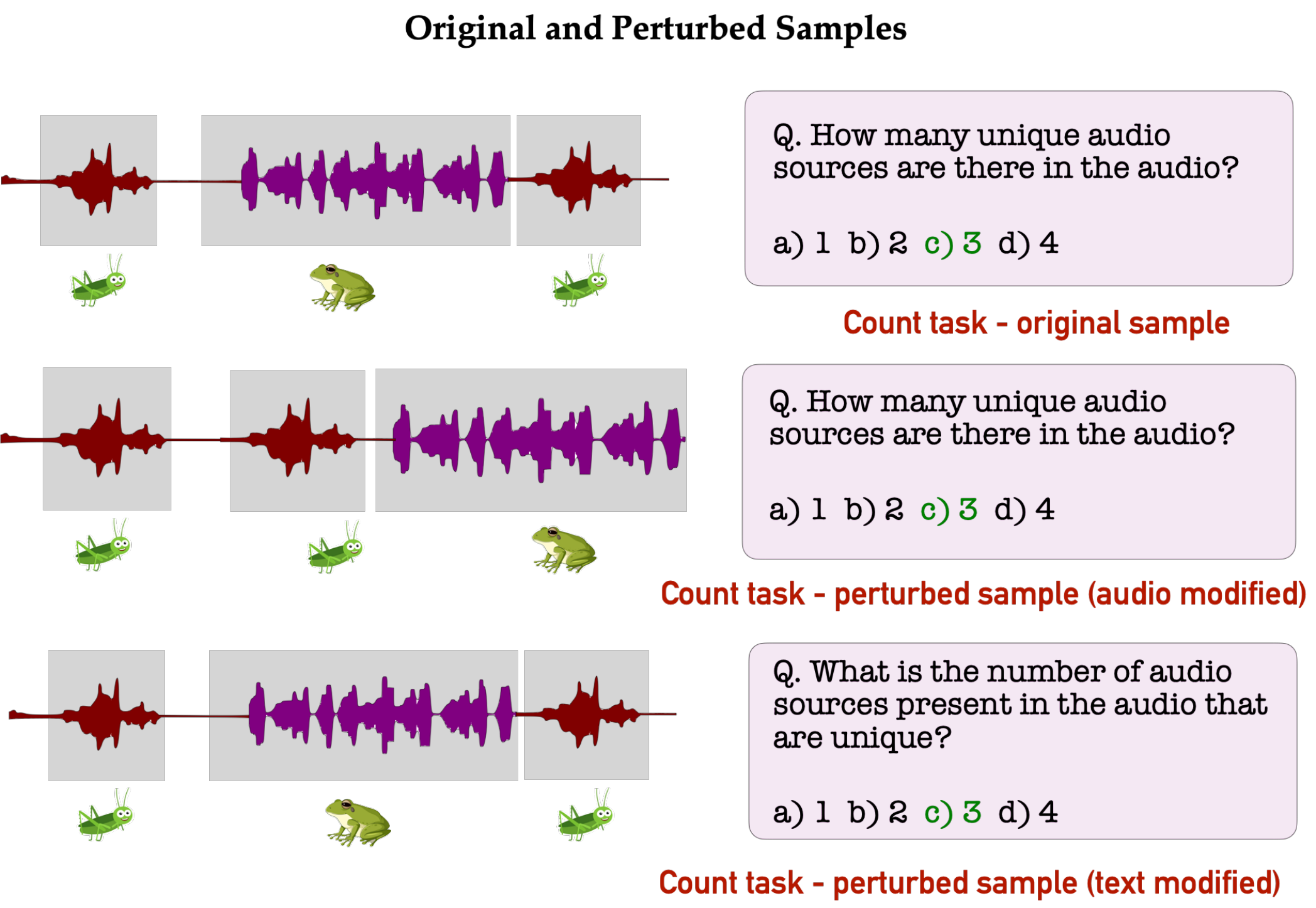}
    \caption{Examples of original sample (from order task), and two perturbed samples, one with audio modification and one with text modification. } 
    \vspace{-0.2in}
    \label{fig:audio-perturbations}
\end{figure}

Examples of data perturbation are shown in Fig,~\ref{fig:audio-perturbations}. 
 The local semantic augmentations are performed in a variety of ways to ensure both diversity (audio and text manipulations) and semantic grounding. For all tasks, the textual modification involves rephrasing of the question in multiple ways while ensuring that the meaning is unchanged (\texttt{aug\_paraphrase()}). For audio perturbations, we use various augmentation strategies.

\begin{itemize}
    \item \texttt{aug\_silence()}: This operation inserts silence (of random duration) at various positions (randomly chosen) between sound events. This augmentation is used in all tasks. 
    \item \texttt{aug\_volume()}: In this setting, the volume of various audio event recordings are amplified or attenuated. This augmentation is used in all tasks. 
    \item \texttt{aug\_shuffle\_order()}: The order of  various audio events are randomized, keeping duration of the audio events unchanged. This augmentation is used in duration task. 

    \item \texttt{aug\_duration()}: The duration of the audio event is varied.  This augmentation is used in the order tasks.

    \item \texttt{aug\_insertion\_deletion()}: The number of audio event repetitions for same class is randomized while combining them. This augmentation is used in counting tasks. 
    
\end{itemize}
For every audio sample, $60$ perturbations are generated (combination of $4$ text modifications and $15$ audio modifications). To limit computation, we chose a subset of $15$ original test samples for each task, leading to a total $2.7k$ augmented samples.

\subsubsection{Uncertainty computation}

Let $\{\x_n^i\}_{i=1}^N$ denote the test-time perturbations corresponding to a sample $\x _n $. For this sample, the uncertainty is measured as,

\begin{equation}\label{eq:uncertainty}
        U_{\x_{n}} = \frac{1}{N} \sum_{i=1}^N {|i : M(\x_n^i) \neq M(\x_n)|}
\end{equation}
where, $M(\x_n)$ is model's prediction (output answer choice) for the test input $\x_n$. As all the samples $i={1..N}$ are generated by preserving the semantics, the uncertainty (Eq.~(\ref{eq:uncertainty})) measures the lack of consistency in the model's response to semantically equivalent input perturbations. An expected uncertainty measure (EUE) is measured as the average of  $U_{\x_{n}} $ for $n=1..K$ test samples. Note that, this measure does not require access to ground truth label.

\subsection{Calibration Error}
For a data point $(\x _n, y_n)\in \mathcal{D}$, if a model makes prediction $\hat{y}$ with confidence $p_n$ (token probability) such that
\begin{equation}
    p_n\propto \text{Pr($\hat{y}_n$ = $y_n$)}
    \label{eqn:calibration_criterion}
\end{equation}
then the model is called as a perfectly confidence-calibrated system. The degree of calibration is often measured using the - expected calibration error (ECE) \cite{jiang2021can}. 

\begin{equation}
    ECE = \sum_{i=1}^{B} w_i * |acc(b_i) - conf(b_i)|
\end{equation}

where $p_n\in [0.0, 1.0]$ is split in $B$ bins $\{b_i\}_{i=1}^{B}$. The $|\cdot|$ is the absolute value operator, bin accuracy $acc(b_i)$ is,
\begin{equation}
    acc(b_i) = \frac{|\hat{y}_n = y_n \,:\, x_n\in \mathcal{D} \,\text{and}\, p_n\in b_i|}{|b_i|}
\end{equation}
and the average bin confidence $conf(b_i)$ is,
\begin{equation}
    conf(b_i) = \frac{\sum \{p_n : p_n\in b_i\}}{|b_i|}
\end{equation}
\subsection{Results and Discussion}
The benchmarking of LALMs using calibration and uncertainty metrics is reported in Table~\ref{tab:calibration-and-uncertainty}. 
 We observe that the accuracy measure does not always provide the full picture with respect to the performance of the LALM model. The combination of SALMONN with LLaMa model gives the best performance in all cases, however, the uncertainty (EUE) and calibration (ECE) errors are lower for the SALMONN-13B model.
 
 This analysis  highlights the need for evaluating LALMs comprehensively to understand if the model's token probabilities are good estimates of the confidence (lower calibration error) and if the model is resilient to semantically grounded perturbations (lower uncertainty error) in addition to the accuracy metrics that are commonly reported. 
 
\section{Summary}
In this paper, we have proposed multiple novel components that help advance the understanding and benchmarking of LALMs. First, the work proposes a novel dataset named temporal reasoning evaluation of audio (TREA) which allows fine-grained assessment of the LALM's capabilities on order, duration and counting of audio events separately. Second, the paper reports their inferiority on these tasks, with the best achieved accuracy being $<50$\% on two out of the three tasks. Finally, the work makes a novel contribution in quantifying the uncertainty of the LALM's predictions at test-time, without requiring the model access or the ground-truth labels.  We hypothesize that the proposals made in this work will shape the next phase of LALM development, where temporal reasoning as well as confidence/uncertainty estimation become important factors in the model design. 

\bibliographystyle{IEEEtran}
\bibliography{mybib}

\end{document}